\documentclass{article}

\usepackage{arxiv}

\usepackage[utf8]{inputenc} 
\usepackage[T1]{fontenc}    
\usepackage{hyperref}       
\usepackage{url}            
\usepackage{booktabs}       
\usepackage{amsfonts}       
\usepackage{nicefrac}       
\usepackage{microtype}      
\usepackage{lipsum}
\usepackage{graphicx}
\usepackage[ruled,linesnumbered]{algorithm2e}
\usepackage{amsmath}
\usepackage{url}
\usepackage{setspace}
\graphicspath{ {./images/} }

\title{Algorithms for Optimizing Fleet Scheduling of Air Ambulances}

\author{
 Joseph Tassone \\
  Department of Computer Science\\
  Lakehead University\\
  Thunder Bay, ON, P7B 5E1 \\
  \texttt{jtasson2@lakeheadu.com} \\
  \And
 Salimur Choudhury \\
  Department of Computer Science\\
  Lakehead University\\
  Thunder Bay, ON, P7B 5E1 \\
}

\begin{document}
\maketitle
\begin{abstract}
Proper scheduling of air assets can be the difference between life and death for a patient. While poor scheduling can be incredibly problematic during hospital transfers, it can be potentially catastrophic in the case of a disaster. These issues are amplified in the case of an air emergency medical service (EMS) system where populations are dispersed, and resources are limited. There are exact methodologies existing for scheduling missions, although actual calculation times can be quite significant given a large enough problem space. For this research, known coordinates of air and health facilities were used in conjunction with a formulated integer linear programming model. This was the programmed through Gurobi so that performance could be compared against custom algorithmic solutions. Two methods were developed, one based on neighbourhood search and the other on Tabu search. While both were able to achieve results quite close to the Gurobi solution, the Tabu search outperformed the former algorithm. Additionally, it was able to do so in a greatly decreased time, with Gurobi actually being unable to resolve to optimal in larger examples. Parallel variations were also developed with the compute unified device architecture (CUDA), though did not improve the timing given the smaller sample size.
\end{abstract}


\section{Introduction}
\label{sec:introduction}
Disasters can strike at any time and in these scenarios it can be devastating if there is not an efficient emergency medical service (EMS) system. While this can be problematic in a city-based environment, it is catastrophic when considering a larger air ambulance system. These vehicles travel over much longer distances, through regions with very dispersed populations. While patient transfers may occur between populous areas, this is not guaranteed during a disaster. Additionally, many air ambulances have a rather small contingency \cite{ornge_2019} which must cover all missions within a certain time frame in order to be effective. It is critical to accurately schedule vehicles in such a way that these potentially vast distances can be managed and patient lives can be secured.

This problem is not new, as there have been a number of exact methodologies suggested in past research \cite{ornge_schedule}. Though solving for optimal is not always unreasonable, since scheduling is NP-hard, it becomes much more time consuming with an increase in the solution space. The issue with this is compounded in a critical system like EMS where time can mean the difference between life and death. Near-optimal metaheuristics provide an effective alternative as they achieve acceptable solutions in a much more manageable window. In this research, known coordinates of air and health facilities were utilized and aided in simulating an actual scheduling scenario. A mission in this case always began at a vehicle occupied base and returned to the same base. Subsets of missions could be performed by each vehicle with a respective mission performing a pickup and delivery before continuing to the next point. Two algorithmic solutions were developed for this problem and each was structured using a sequential and parallel methodology. For the latter, the Compute Unified Device Architecture (CUDA) platform was employed and timed against the former for each algorithm. Overall, the goal was to develop a model with integer linear programming and then formulate algorithms which could attain a similar solution to the optimal. 

This paper is arranged as follows. Section \ref{sec:related_work} describes the related work with an a description of past techniques and solutions. Section \ref{sec:model} outlines the modelling of the problem, emphasizing constraints and the objective. Section \ref{sec:algorithm} displays the algorithmic solutions, describing the neighbourhood and tabu search, as well as the initialization technique. Sections \ref{sec:results} and \ref{sec:conclusion} respectively show the results and conclude the paper.

\section{Related Work}
\label{sec:related_work}
For scheduling problems, neighbourhood searches have been well-documented in achieving successful results. Da\u{g}layan and Karakaya suggested a solution for scheduling ambulances following a disaster and formulated it as a capacitated vehicle routing problem (CVRP) \cite{karakaya_2016}. In their research, they minimized the routes and then reduced the average travel time to health facilities. For this process they developed both a genetic algorithm and nearest-neighbour heuristic. Both operated independently and were compared in three different test cases. These were based on capacity where light damage had 1 victim, medium damage could have 1 to 8 victims, and heavy damage had beyond 8. While the authors claimed that further study was required, when compared against CVRP benchmarks the genetic algorithm achieved shorter distances. This work does require additional testing, as there is a limitation in utilizing a single facility and ambulance.

Tabu search is another successful algorithm in scheduling, being an extension of the neighbourhood-based algorithms. In this type of solution recently explored neighbourhoods which improved the result are ignored, preventing the risk of achieving only a local optimum \cite{EDELKAMP2012633}. Oberscheider and Hirsch utilized this methodology in conjunction with real-patient data to ensure the efficient transport of patients for lower Austria's Red Cross \cite{10.1186/s12913-016-1727-5}. They developed an algorithm based on the static multi-depot heterogeneous dial-a-ride problem and generated all possible combinations of patient deliveries considering a set of constraints in order to solve it. For each prior generation, set partitioning was performed to achieve an initial solution and then these were inputted into a Tabu search metaheuristic for optimization. The authors claim that the work is an improvement upon previous variations as it considers non-static service times that depended on the combination of patients, their transport mode, the vehicle type, and the pickup or delivery locations.

Another research presented by Repoussis et al. formulated a mixed integer linear programming model for the scheduling and routing of large casualty disasters \cite{REPOUSSIS2016531}. They sought to use a minimized amount of resources while also being able to achieve a reduced response time. Their solution was developed as a hybrid multi-start local search, involving an exact construction heuristic followed by an optimization by iterated Tabu search. Initialization utilized a greedy technique for assignments and then the problem in its reduced state was optimized. A high-quality upper bounds was achieved and these were inputted into the Tabu search, repeating until the stopping condition was met.

Most solutions involve sequential designs where a solution must iteratively solve until reaching an optimal. While not a new trend, parallelization has been utilized to achieve higher calculation speeds through the graphics processing unit \cite{LEE2012175}. The algorithms must be redeveloped to take advantage of the multiple simultaneous threads and loops are replaced by these to perform calculations. Additionally, threads are divided into blocks which can be an issue in larger problems. In this case thread communication occurs within blocks, but synchronization cannot happen between them. For these scenarios block level design must be considered along with thread design. Regardless, the primary bottleneck is related to the kernel call to GPU when the problem space is small enough. Race conditions are another potential issue, though CUDA provides both shared memory and locks to deal with this problem \cite{cuda_toolkit_documentation_2019}.

While prior parallel research has been performed, it is limited in terms of optimization problems of this type. Schulz et al. performed a survey revealing a small amount of literature on applying GPU parallelization to local or Tabu search \cite{Schulz2013}. The same can be said when applied to regional or air ambulance problems. Hussai et al. redeveloped the particle swarm optimization algorithm by using CUDA to perform partially coalescing memory accesses \cite{7829615}. When compared against benchmarks the new algorithm was able to gain a significant improvement in time compared to a similar sequential variation. Similarly, Fabris and Krholing used benchmarks and appleid CUDA to a co-evolutionary differential evolution algorithm for solving min-max optimization problems \cite{FABRIS201210324}. Their research showed that the algorithm was able to converge to a near optimal result, although in a much better time when scaled up.

\section{Model}
\label{sec:model}
Scheduling of air ambulances follows predetermined positions, where ambulances are already placed at bases. Time is also a factor as scheduling relies on patients arriving at certain periods. This may be due to urgency from an incident or transfers for more specialized care. While there are potentially a large number of bases in an EMS system, this scheduling problem only deals with the subset of bases that have vehicles assigned to them. In general, unless otherwise specified, the scheduling problem treats the terms vehicles and bases as synonymous.

\begin{figure}[!h]
\centering
  \includegraphics[width=4.0in]{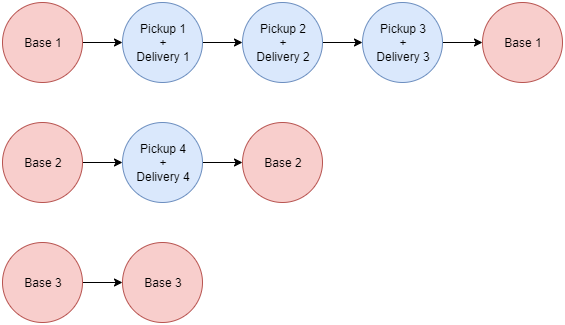}
\caption{Basic scheduling setup for missions and bases.}
\label{fig_pickup_delivery}
\end{figure}

In this case a mission starts at a base and can be made up of at least a single pickup and delivery. However, rather than a simple return to base, another mission may be assigned to that vehicle, where it will continue from the previous delivery to next pickup. Therefore, the schedule will consist of each base being assigned a set of missions, followed by a return to the original base. An example of this problem and all possible setups can be seen in Figure \ref{fig_pickup_delivery}. There are three bases, with four missions. So long as the constraints are met, each base can have either as many missions as possible assigned or none at all.

\medbreak

\noindent The potential bases consist of the following two sets:
\\$P$: the set of all fixed-wing plane occupied bases with each being a 4-tuple of form

$a_k=<k\: s\: \phi_{k}\:\psi_{k}>,a_k \in P \:\forall \:k= 1, ..., 4$

$k \equiv$ base ID

$s \equiv$ vehicle speed

$\phi_{k} \equiv$ row coordinate of base $k$

$\psi_{k} \equiv$ column coordinate of base $k$
\\$H$: the set of rotary-wing helicopter occupied bases with each being a 4-tuple of form

$h_k=<k\: s\: \phi_{k}\:\psi_{k}>,h_k \in H \:\forall \:k = 5, ...,12$

$k \equiv$ base ID

$s \equiv$ vehicle speed

$\phi_{k} \equiv$ row coordinate of base $k$

$\psi_{k} \equiv$ column coordinate of base $k$

\medbreak

\noindent The set set of all missions consists of:
\\$M$: the set of all missions, with each being a 6-tuple of form

$m_n=<n\:\phi_{p}\:\psi_{p}\:\phi_{d}\:\psi_{d}\:\rho>,m_n \in M \:\forall \:n$

$n \equiv$ mission ID

$\phi_{p} \equiv$ row coordinate of patient pick-up location

$\psi_{p} \equiv$ column coordinate of patient pick-up location

$\phi_{d} \equiv$ row coordinate of patient delivery location

$\psi_{d} \equiv$ column coordinate of patient delivery location

$
\rho =
\begin{cases}
  1 & \text{if mission requires rotary-wing helicopter}\\  
   0 &\text{otherwise}   
\end{cases}
$

\medbreak

Sets $H$ and $P$ are vehicle occupied bases, where both are constant and locked to their specific coordinates. No other bases can be utilized and following a set of missions there must be a return to the original base. Both sets are parts of the full set of bases which can be utilized, stated as $P \cup H = K$. Additionally, rather than total distance, the travel time from one position to another is considered. Helicopters and planes travel at different speeds, meaning that regardless of distances they will arrive at different times. Therefore, speed must considered for each set in order to perform accurate scheduling. Missions have specific requirements where only a helicopter occupied base may perform certain missions.

There are a set of variables for the problem, with the first being a binary $x_{ijk}$. A respective instance will be 1 if a mission $i$ to $j$ are connected and starting from base $k$. Additionally, the variable $u_i$ is utilized for subtour elimination based on the Miller-Tucker-Zemlin constraint \cite{DESROCHERS199127}. For the exclusion of subtours the value can be $i=\{1,...,n\}$ for each base $k$. The model also considers a set of constants $p$, $w_n$, $b_k$, and $f_n$. Both $p$ and $w_n$ contribute to time constraints where $p$ is the maximum flight time allowed by a vehicle in a day (10 hours for the purposes of this research) and $w_n$ is the time by which a vehicle must arrive at a base by. In the scheduling problem they are independent as once a vehicle arrives for a mission it must wait until its respective $w_n$ to continue to the next point. The constant $p$ only considers actual time in flight and does not recognize waiting. Both $b_k$ and $f_n$ are utilized for limiting certain bases to certain missions. $b_k$ states the type of vehicle that is at an individual base (0 for helicopter and 1 for plane), while $f_n$ describes the vehicle requirement that must be used for a mission (0 for helicopter-only and 1 for any vehicle). 

In addition to distance calculations, the respective travel time is considered as determined by the speed of the vehicle. Based on known statistics about air ambulance vehicles, rotary-wing helicopters are assigned a speed of 300km/h \cite{pilatus_aircraft_ltd_2019} and fixed-wing planes are assigned a speed of 500km/h \cite{leonardo}. The matrix $d$ is the result of the calculation, where the distance from $i$ to $j$ is determined and divided by the speed of the vehicle. The respective result is then placed into $l$ where $l_0$ is for helicopters and $l_1$ is for planes. The distance is determined for a base as the sum of each mission's patient pick-up and delivery locations. Since this is not simulated data, Haversine distance is considered. The full calculation can be completed with the following:

\begin{equation}
  \label{e1}
  \resizebox{0.9\columnwidth}{!}{
    $\mathit{d_{ijl}} =
    \frac{2r\sin^{\:\!\!-1}\left(
    \sqrt{\sin^{\:\!\!2} \left(\frac{ \phi - \phi_{p} }{2}\right) +  \cos(\phi)\cos(\phi_{p})\sin^{\:\!\!2} \left(\frac{\psi - \psi_{p}}{2}\right)}\right)}{K_s}$
  }
\end{equation}

In the above formula $\phi$ represents row locations, while $\psi$ applies to column location values. $K_s$ is used to indicate a specific speed of a vehicle at a base, which is divided by the Haversine distance calculated. The resulting matrix is of dimension $(n+k) \times (n+k) \times 2$ and referenced for achieving total travel time for each mission. The optimization model takes the following form:

\begin{align} 
&\textbf{minimize}\nonumber\\
&\sum_{i}\sum_{j}\sum_{k}x_{ijk}d_{ijl}; \:l = b_k
\label{obj_s}
\\
&\textbf{subject\: to}\nonumber\\
&\sum_{i}\sum_{k}x_{ijk} = 1 \:; \forall \:j
\label{con1_s}\\
&\sum_{j}\sum_{k}x_{ijk} = 1 \:; \forall \:i
\label{con2_s}\\
&\sum_{i}x_{ink} = \sum_{j}x_{njk} \:; \forall \:k, n
\label{con3_s}\\
&\sum_{i}x_{ijk} \leq 1 \:; \forall \:j, k
\label{con4_s}\\
&\sum_{j}x_{ijk} \leq 1 \:; \forall \:i, k
\label{con5_s}\\
&\sum_{i}\sum_{j}x_{ijk}d_{ijl} \leq p \:; \forall \:k, l = b_k
\label{con6_s}\\
&x_{ijk}d_{ijl} \leq x_{ijk}w_{j}\:; \forall\:j, k, l = b_k , i \leq n
\label{con7_s}\\
&x_{ijk}(d_{ijl}w_{i}) \leq x_{ijk}w_{j}\:; \forall\:j, k, l = b_k , i > n
\label{con8_s}\\
&x_{ijk}(b_{k} - f_{i}) \leq 0\:;\forall \:i, j, k
\label{con9_s}\\
&u_{i} - u_{j} + n(x_{ijk}) \leq n - 1 \:; 1 \leq i \neq j \leq n, \forall \:k
\label{con10_s}\\
&x_{ijk} = 0\:; \:(i\neq j, k) \in K
\label{con11_s}\\
&x_{ijk} \in \{1,0\}\:; \forall \:i, j, k
\label{con12_s}\\
&u_{i} \in \{i=1,..,n\}
\label{con13_s}
\end{align} 

Equation \ref{obj_s} describes the objective function, where the total time flown by each vehicle to its assigned missions are minimized. Each set's $d_{ijl}$ is summed based on its corresponding $x_{ijk}$ variables which have a 1. The actual distance is determined by the constant $b_k$, where the index of $l$ equals the respective value. The objective function is also constrained based on the rules set by Equations \ref{con1_s} to \ref{con13_s}. 

Each mission can only be completed once, constrained by Equations \ref{con1_s} and \ref{con2_s}. Equation \ref{con3_s} guarantees route continuity, where each mission must continue from one to the other along their assigned route. To ensure that missions leave and return to the same base, the model considers Equations \ref{con4_s} and \ref{con5_s}. As previously stated, vehicles can only fly so long which determined by $p$. The constraint for this is upheld by Equation \ref{con6_s} for each base $k$ set of missions. The remaining time constraints are determined by Equations \ref{con7_s} and \ref{con8_s}. The first is for when a vehicle is travelling from a base to a mission. In this case the base does not have a time limit and begins at 0. The second considers either a mission to a mission, or a mission returning to a base. In the latter case the limit is usually assumed to be 24, which corresponds to the number of hours in a day. Regardless of travel time the vehicle must wait until that missions time limit to travel to the next one. The distance summed with the previous time limit must be less than the next mission's time limit. 

To force each mission to be correctly assigned to a valid base, the model uses Equation \ref{con9_s}. There is only one invalid permutation for this assignment, where a plane is assigned to a helicopter-only mission. This is prevented as all other possibilities are forced to be less than or equal to 0. The subtour elimination constraint is outlined by Equation \ref{con10_s} and uses the Miller-Tucker-Zemlin formulation. For this constraint, $u_i$ is a variable associated with each mission and is used to eliminate sets that do not begin and end at a base. The variable is equal to the order in which each mission occurs for each base $k$, where the base begins at the first index \cite{doi:10.1080/13675567.2019.1630374}. To further limit the formulation, Equation \ref{con10_s} prevents bases from travelling to other bases other than themselves. Lastly, Equations \ref{con11_s} and \ref{con12_s} limit the values which can be assigned to each variable.

\clearpage

\section{Algorithms}
\label{sec:algorithm}
All the solutions within this section were designed with consideration of the constraints and variables in the scheduling model above. The algorithms each had a sequential and parallel CUDA implementation. As there were only minor changes within the algorithms, Algorithm \ref{local_search_scheduling} and \ref{tabu_search_scheduling} are outlined as the sequential versions. Any other alterations are given following a description of each one.

\subsection{Scheduling Initialization Algorithm}
Due to the constraints, there is no guarantee that a randomized initialization will be able to generate a valid solution. Since vehicles must fly under a certain period and within a particular limit there is a increased difficulty in discovering a starting state. Therefore, Algorithm \ref{schedule_initialize} was developed to find an initialized organization for optimization. To clarify, the algorithm is not designed to find a near-optimal solution, only to determine the existence of any solution. Essentially, the method will attempt to assign missions to bases and upon failure will perform swaps. This will improve the current result and may increase the chances of finding a valid permutation. If after so many attempts it does not find a solution it will report an error.

The algorithm accepts a set of occupied bases and a set of mission data as its input. Lines 1-4 show the constant data that must be abided by including the distance matrices for each respective vehicle type, the mission times determining when a vehicle must arrive by, and the daily flight limit for a vehicle. The $Solution$ matrix declared at line 5 will hold each mission with a row starting and ending with a base, and compatible missions will then fill in between these indices. The remainder of the algorithm is broken up into a section for helicopter-only mission assignments (lines 6- 42) and then a section for the remaining missions (lines 43-77). For both, the outer loops will only terminate once all are assigned. 

The first step at line 8 is to determine the unassigned mission with the smallest time limit. There is an urgency for this to be completed first as the time limit constraint cannot be broken. As per line 9, every base will be checked and the current best fit will be held in $CurrentMin$. The next set of lines follow the constraints set by the model, limiting which bases can accept a mission. If a base breaks any of the requirements, then the algorithm continues to the next possible base. Lines 10-12 ensure that only a helicopter occupied base may accept a helicopter-only mission. Lines 13-34 are the time limit constraints, ensuring that vehicle does not pass the arrival time set for a respective mission. This part is broken into two parts based on whether a base already has missions or has yet to have one assigned to it. As the mission is being placed between two indicies it must consider the vehicle arriving at the new mission and the travel new mission to the next one.
Additionally, once a vehicle arrives at a mission it must wait until that time limit to move on to the following. If it is starting at a base the limit is 0 and not considered, while if it from another mission it must utilize the distance to the next plus the previous time limit. If the vehicle is returning to a base then the time cannot pass 24, which is represents the number of hours in the day. Lines 29-31 performs a check on the current flight time of each set of missions. This constraint only considers actual flight time and not the waiting enforced by the previous constraints. Once these restrictions have been met, lines 32-34 check if the new assignment is an improvement over the prior. The result is saved if there has yet to be an update to $CurrentMin$ or if the new value is smaller than the previous.

If $CurrentMin$ has a solution then it is added to the $Solution$ matrix at line 40. There is the possibility that following the loop there will be no available positions to place the mission. If this occurs then lines 36-37 attempt a reorganization in order to find a new valid permutation. These swap are performed by Algorithm \ref{local_search_scheduling} and discussed below. The algorithm returns an error if no reorganization is possible, no missions have been assigned yet, or following a reorganization the assignment fails.

Lines 43-77 perform nearly the exact same process, with the exception that there is no mission compatibility requirement. In order to validate the correct matrix to choose from, the algorithm includes a flag called $VehicleType$ (line 47). Each remaining check performs almost identically to the last section; however, the flag is used to ensure the corresponding distance matrix matches the vehicle at each base.

The addition of parallelism to this algorithm was simplistic, as race conditions were only an issue for $CurrentMin$. CUDA operates through concurrent thread organization, so the loops at lines 7 and 44 could instead be replaced with their own threads. The continues are also replaced as each thread controls its own base. This allowed the $MinIndex$ to be checked simultaneously for each base. A mutex check was performed at line 32 to add to $CurrentMin$ along with the thread ID. At line 40, if the thread ID matched the $CurrentMin$, then it updated it with its solution. If there was currently no updated solution, a thread would be allocated to performing the swaps.

\begin{algorithm}[b!]
\scriptsize
 \KwData{Base, mission pickup, and mission destination data}
 \KwResult{Assignment of missions to bases}
 $HeliDistances \leftarrow$\ distance matrix helicopter bases to each respective mission and each mission to each other mission\;
 $PlaneDistances \leftarrow$\ distance matrix plane bases to each respective mission and each mission to each other mission\;
 $MissionTimes \leftarrow$\ times by which a vehicle must arrive at a particular mission delivery point by\;
 $FlightLimit \leftarrow$\ an integer stating the amount of allowed flight time per day\;
 $Solution \leftarrow$\ a matrix initialized with each base in the first and final index for each row\;
 \While{all helicopter only missions are not assigned}{
    \For{$i \gets 1$ to number of bases}   {
        $MinIndex \leftarrow$\ current unassigned helicopter only mission with the smallest value in $MissionTimes$\;
        $CurrentMin \leftarrow$\ current best $i$ base for $MinIndex$\;
        
        \If{base $i$ does not contain a helicopter}{
           Continue to the next $i$ base\;
        }
        \If{base $i$ does not have any missions}{
            \If{$HeliDistances$ from base $i$ to $MinIndex > MissionTimes$ of $MinIndex$} {
                Continue to the next $i$ base\;
            }
            \If{$MissionTimes$ of $MinIndex$ + $HeliDistances$ from $MinIndex$ to base $i > 24$} {
                Continue to the next $i$ base\;
            }
        }
        \ElseIf{base $i$ does have missions assigned to it}{
            \If{$MissionTimes$ of previous assigned base $i$ mission $+ HeliDistances$ from previous assigned base $i$ mission to $MinIndex > MissionTimes$ of $MinIndex$} {
                Continue to the next $i$ base\;
            } 
            \If{$MissionTimes$ of $MinIndex$ + $HeliDistances$ from $MinIndex$ to base $i > 24$} {
                Continue to the next $i$ base\;
            }
        }
        \If{assigning $MinIndex$ mission to base $i$ exceeds $FlightLimit$}{
           Continue to the next $i$ base\;
        }
        \If{$CurrentMin$ is empty or the new $i$ gives a better result}{
           Update $CurrentMin$\;
        }
    }
    \If{$CurrentMin$ is empty}{
        Perform mission swaps from Algorithm \ref{local_search_scheduling} and retry for new arrangement
    }
    \Else{
        Update $Solution$  with $CurrentMin$ values\;
    }
 }
 \caption{Scheduling Initialization Algorithm}
 \normalsize
 \label{schedule_initialize}
\end{algorithm}

\pagebreak

\begin{algorithm}[!t]
\setcounter{AlgoLine}{42}
\scriptsize
\While{all remaining missions are not assigned}{
    \For{$i \gets 1$ to number of bases}   {
        $MinIndex \leftarrow$\ current unassigned mission with the smallest value in $MissionTimes$\;
        $CurrentMin \leftarrow$\ current best $i$ base for $MinIndex$\;
        $VehicleType \leftarrow$\ flag determining whether to select from $HeliDistances$ or $PlaneDistances$\;

        \If{base $i$ does not have any missions}{
            \If{$VehicleType$ type from base $i$ to $MinIndex > MissionTimes$ of $MinIndex$} {
                Continue to the next $i$ base\;
            }
            \If{$MissionTimes$ of $MinIndex + VehicleType$ type from $MinIndex$ to base $i > 24$} {
                Continue to the next $i$ base\;
            }
        }
        \ElseIf{base $i$ does have missions assigned to it}{
            \If{$MissionTimes$ of previous assigned base $i$ mission $+ VehicleType$ type from previous assigned base $i$ mission to $MinIndex > MissionTimes$ of $MinIndex$} {
                Continue to the next $i$ base\;
            } 
            \If{$MissionTimes$ of $MinIndex + VehicleType$ type from $MinIndex$ to base $i > 24$} {
                Continue to the next $i$ base\;
            }
        }
        \If{assigning $MinIndex$ mission to base $i$ exceeds $FlightLimit$}{
           Continue to the next $i$ base\;
        }
        \If{$CurrentMin$ is empty or the new $i$ gives a better result}{
           Update $CurrentMin$\;
        }
    }
    \If{$CurrentMin$ is empty}{
        Perform mission swaps from Algorithm \ref{local_search_scheduling} and retry for new arrangement\;
    }
    \Else{
        Update $Solution$ with $CurrentMin$ values\;
    }
 }
  \normalsize
\end{algorithm}

\subsection{Neighbourhood Search Fleet Scheduling Optimization}
Algorithm \ref{local_search_scheduling} accepts either the initialized or the partially complete data from Algorithm \ref{schedule_initialize}. The goal of the algorithm is to find an organization of missions and bases which minimizes the total travel time. As already discussed, the design of the previous algorithm was not to find an optimal solution, only a valid one. Conversely, Algorithm \ref{local_search_scheduling} was formulated to be performed through multiple instances in order to minimize the total time to near-optimal. Though the initialization method uses this algorithm, it only performs a single iteration to attempt a new result. In order to ensure that Algorithm \ref{local_search_scheduling} has a diverse set of possibilities, the loops at lines 6 and 7 may be exchanged with random index permutation vectors.

The algorithm functions by moving missions to new bases, essentially having another vehicle take responsibility for that mission. It will only end once no further improvement is found. If utilizing a set of permutation vectors, this means no better result is found in the current permutation and provides the opportunity for another iteration. A similar set of constraints from the previous algorithm are used and updates are performed to the $Solution$ matrix. There are four sets of loops with line 6 allowing for looping through the bases and line 7 for looping through the respective missions assigned to a base. The loops at lines 9 and 17 perform a similar function, though they are the corresponding positions being move to.

Similar to Algorithm \ref{schedule_initialize}, it includes a $CurrentMin$ variable for keeping track of the best current position for swapping. Once every valid combination is checked, the $CurrentMin$ updates the $Solution$ at line 43. Swapping within a base is restricted by lines 10-12, as the current order is already constrained by the time. Additionally, certain planes cannot accept helicopter-only missions as stated in lines 13-15. There is a $VehicleType$ flag for determining which matrix to select from for the remainder of the algorithm. This is based on the specific vehicle that is assigned.

\begin{figure}[!h]
\centering
  \includegraphics[width=3.0in]{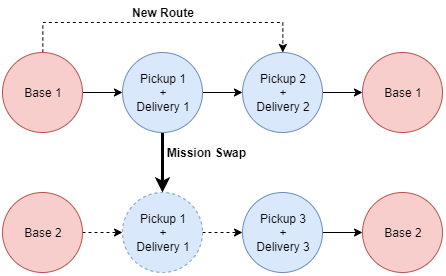}
\caption{The movement of one mission to another base.}
\label{fig_mission_swap}
\end{figure}

An example of the process can be seen in Figure \ref{fig_mission_swap}, where the first mission connected to base 1 is moved to base 2. Once a swap occurs base 1 connects directly to mission 2, while the base 2 setup adds two new links between the new position of mission 1. During the movement the new position of the mission cannot force the vehicle's flight limit to be exceed, expressed by lines 18-20. Similarly, the new links must not exceed any of the time limits set by the lines 21-36. This function is identical to the design of the previous algorithm where the check depends on where the mission is being moved to. Again a vehicle must wait until the current mission's time limit before proceeding, meaning each proceeding time limit must be considered in the check. If the movement passes these requirements and is the current best option, then it is added to $CurrentMin$ at line 38.

Like the previous algorithm, this one can be made easily parallel by the removal of the loop at line 9. This allows every base to be simultaneously checked and for the $CurrentMin$ to be updated with the best result quicker. The outer loop at line 6 could have instead been replaced, though this would have given a different value than the original, making accurate comparison difficult. Either design can be run multiple times and so long as an index ordering is done the solution may be improved.

\begin{algorithm}[!t]
\scriptsize
 \KwData{Initial Solution Data From Algorithm \ref{schedule_initialize}}
 \KwResult{Assignment of missions to bases}
 $HeliDistances \leftarrow$\ distance matrix helicopter bases to each respective mission and each mission to each other mission\;
 $PlaneDistances \leftarrow$\ distance matrix plane bases to each respective mission and each mission to each other mission\;
 $MissionTimes \leftarrow$\ times by which a vehicle must arrive at a particular mission delivery point by\;
 $FlightLimit \leftarrow$\ an integer stating the amount of allowed flight time per day\;
 \While{$improvement$}{
    \For{$i \gets 1$ to number of bases}{
        \For{$j \gets 1$ to number of $i$ assigned missions}   {
           $CurrentMin \leftarrow$\ current best swap position\;
           \For{$k \gets 1$ to number of bases}   {
                \If{$i = k$}{
                    Continue to next $k$\;
                }
                \If{vehicle at base $k$ is not compatible with mission $j$}{
                    Continue to next $k$\;
                }
                $VehicleType \leftarrow$\ flag determining whether to select from $HeliDistances$ or $PlaneDistances$\;
                \For{$l \gets 1$ to number of $k$ assigned missions} {
                    \If{assigning $j$ mission to base $k$ exceeds $FlightLimit$}{
                       Continue to the next $k$ base\;
                    }
                    \If{base $k$ does not have any missions}{
                        \If{$VehicleType$ type from base $k$ to mission $j > MissionTimes$ of $j$} {
                            Continue to the next $k$ base\;
                        }
                        \If{$MissionTimes$ of $j + VehicleType$ type from $j$ to base $k > 24$} {
                            Continue to the next $k$ base\;
                        }
                    }
                    \ElseIf{base $k$ does have missions assigned to it}{
                        \If{$MissionTimes$ mission $l + VehicleType$ type of $l$ to $j > MissionTimes$ of $j$} {
                            Continue to the next $k$ base\;
                        } 
                        \If{$MissionTimes$ of $j + VehicleType$ type from $j$ to mission after $l > MissionTimes$ of mission after $l$} {
                            Continue to the next $k$ base\;
                        }
                    }
                    \If{$CurrentMin$ is empty or the new assignment gives a better result}{
                       Update $CurrentMin$\;
                    }
                }
           }
           \If{$CurrentMin$ is not empty}{
            Update $Solution$ with result in $CurrentMin$\;
           }
        }
    }
 }
 \caption{Neighbourhood Search Fleet Scheduling}
 \label{local_search_scheduling}
 \normalsize
\end{algorithm}


\subsection{Tabu Search Fleet Scheduling Optimization}
This algorithm functions as a modification of Algorithm \ref{local_search_scheduling}, with a near identical design. The primary differences are the variables included at lines 5 and 6. The $TabuList$ prevents recently explored neighbourhoods from being explored again. There is a chance that the algorithm may become locked into a local minimum, preventing optimum exploration of the space. Therefore, the $TabuList$ helps to prevent this, keeping certain regions restricted for a number of iterations based on the $TabuCounter$.

During the iterative process if a region which has recently been explored is already in the list, then it is skipped as per lines 20-27. The respective counter for each value is then reduced at the end of the current iteration with line 56. Additionally, a value is added to $TabuList$ only if improves the value determined by lines 52-55. If permutation vectors are used for the outter loops then improvement may be discovered through multiple runs. In this case the $TabuList$ variable is maintained through proceeding iterations.

The parallel implementation is identical to Algorithm \ref{local_search_scheduling}. The inner loop at line 11 is replaced and continues are removed as each base receives its own thread. Also similar to the previous, Algorithm \ref{tabu_search_scheduling} will result in the exact same answer as the sequential variation. The updates are still handled by a single thread, which is determined by the best result in $CurrentMin$.

\medbreak

\begin{algorithm}[!h]
\scriptsize
 \KwData{Initial Solution Data From Algorithm \ref{schedule_initialize}}
 \KwResult{Assignment of missions to bases}
 $HeliDistances \leftarrow$\ distance matrix helicopter bases to each respective mission and each mission to each other mission\;
 $PlaneDistances \leftarrow$\ distance matrix plane bases to each respective mission and each mission to each other mission\;
 $MissionTimes \leftarrow$\ times by which a vehicle must arrive at a particular mission delivery point by\;
 $FlightLimit \leftarrow$\ an integer stating the amount of allowed flight time per day\;
 $TabuList \leftarrow$\ list of recently explored neighbourhoods\;
 $TabuCounter\leftarrow$\ how long a neighbourhood can be held within a list\;
 \While{$improvement$}{
    \For{$i \gets 1$ to number of bases}{
        \For{$j \gets 1$ to number of $i$ assigned missions}   {
           $CurrentMin \leftarrow$\ current best swap position\;
           \For{$k \gets 1$ to number of bases}   {
                \If{$i = k$}{
                    Continue to next $k$\;
                }
                \If{vehicle at base $k$ is not compatible with mission $j$}{
                    Continue to next $k$\;
                }
                $VehicleType \leftarrow$\ flag determining whether to select from $HeliDistances$ or $PlaneDistances$\;
                \For{$l \gets 1$ to number of $k$ assigned missions} {
                    \If{Selection $k$ or $l$ are in the $TabuList$}{
                        \If{Selection is of an index $k$} {
                            Continue to next $k$ index\;
                        }
                        \Else{
                            Continue to next $l$ index\;
                        }
                    }
                
                    \If{assigning $j$ mission to base $k$ exceeds $FlightLimit$}{
                       Continue to the next $k$ base\;
                    }
                    \If{base $k$ does not have any missions}{
                        \If{$VehicleType$ type from base $k$ to mission $j > MissionTimes$ of $j$} {
                            Continue to the next $k$ base\;
                        }
                        \If{$MissionTimes$ of $j + VehicleType$ type from $j$ to base $k > 24$} {
                            Continue to the next $k$ base\;
                        }
                    }
                    \ElseIf{base $k$ does have missions assigned to it}{
                        \If{$MissionTimes$ mission $l + VehicleType$ type of $l$ to $j > MissionTimes$ of $j$} {
                            Continue to the next $k$ base\;
                        } 
                        \If{$MissionTimes$ of $j + VehicleType$ type from $j$ to mission after $l > MissionTimes$ of mission after $l$} {
                            Continue to the next $k$ base\;
                        }
                    }
                    \If{$CurrentMin$ is empty or the new assignment gives a better result}{
                       Update $CurrentMin$\;
                    }
                }
           }
           \If{$CurrentMin$ is not empty}{
            Update $Solution$ with result in $CurrentMin$\;
            Add selection to the $TabuList$ and initiate $TabuCounter$ for the selection\;
           }
           Reduce the $TabuCounter$ for each within the $TabuList$\;
        }
    }
 }
 \caption{Tabu Search Fleet Scheduling}
 \label{tabu_search_scheduling}
 \normalsize
\end{algorithm}

\section{Results}
\label{sec:results}
Eight datasets were employed for testing the algorithms, all being based on known coordinates of air and health facilities. The Numba library \cite{numba_2018} was applied in order to speed up the programs through the JIT compiler. All programs, including Gurobi scripts, were tested with a system containing 128 GBs of RAM with a 24 core Intel Core i9-7920X X-series processor. CUDA versions made use of a GeForce GTX 1080 Ti graphics card as well as the prior technologies mentioned.

\begin{figure}[]
\centering
  \includegraphics[width=5.5in]{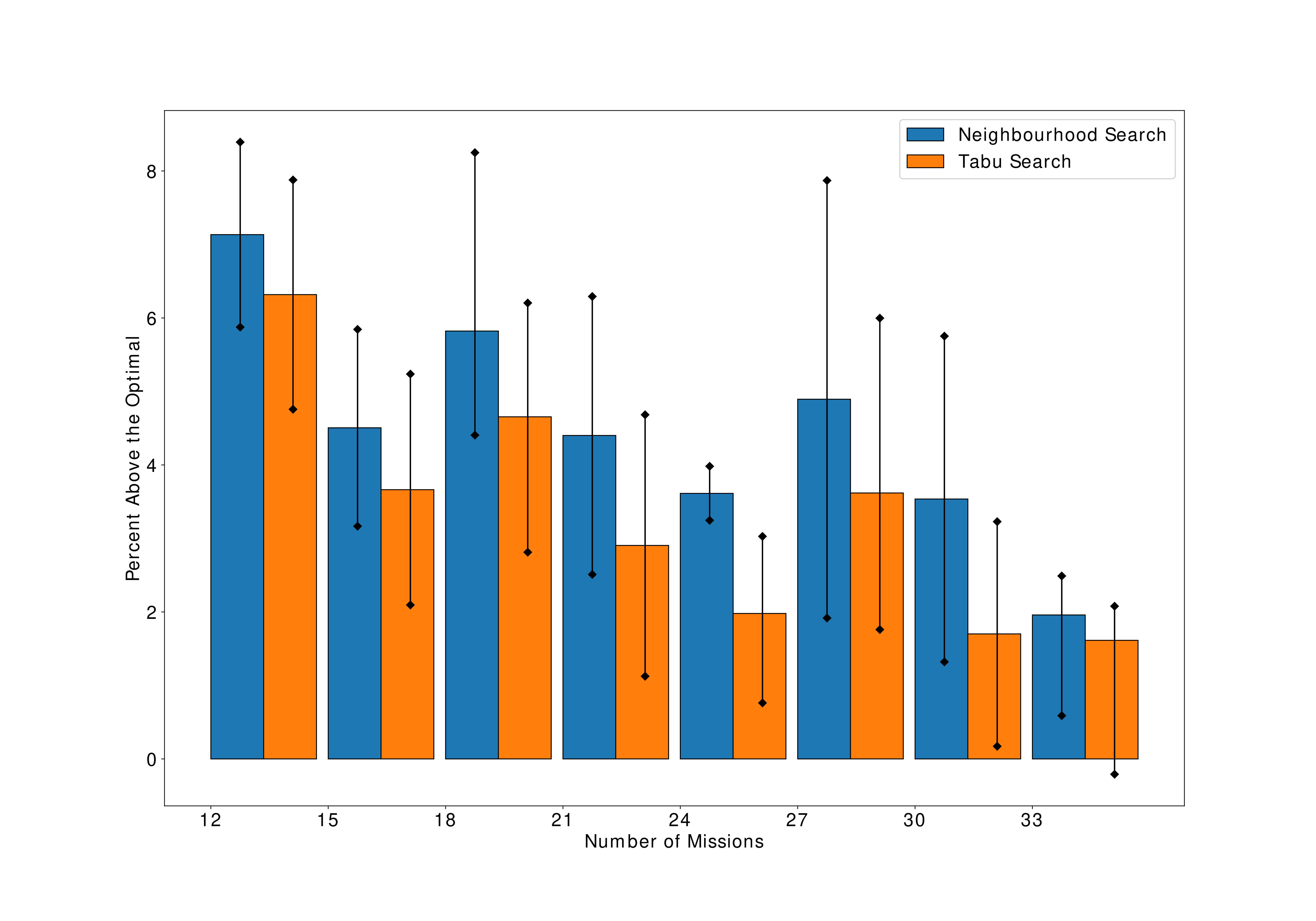}
\caption{Comparison of scheduling algorithms against the optimized solution.}
\label{fig_results_scheduling}
\end{figure}

\begin{figure}[]
\centering
  \includegraphics[width=5.5in]{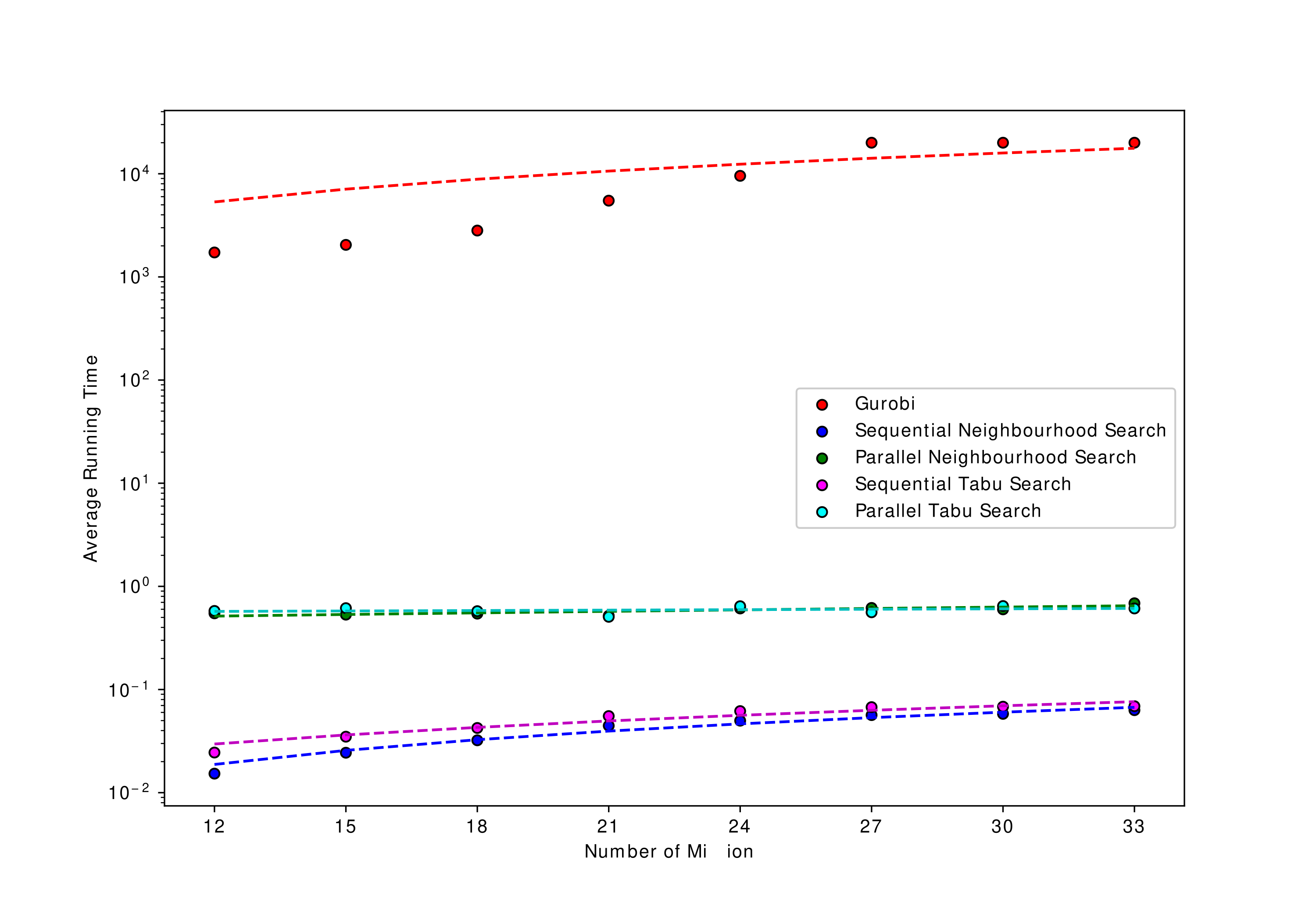}
\caption{Comparison of scheduling runtimes.}
\label{fig_results_time_scheduling}
\end{figure}

Each algorithm ran 10 times, with all results being summarized in Figures \ref{fig_results_scheduling} to \ref{fig_results_time_scheduling} and Tables \ref{fig_times_table_scheduling} to \ref{fig_results_table_scheduling}. A subset of each set of missions were chosen and divided from 12 to 33. 12 vehicles were used consisting of 8 rotary-wing helicopters and 4 fixed-wing planes. The vehicle's locations were locked to the bases and at the end of a set of missions each vehicle needed to return to the same base. Time limits were randomly generated for each, needing to be completed within a 24 hour period. This factor in addition, to a 10 hour flight limit, prevented sets from going over 33 missions. While under the right circumstances it is possible for vehicles to complete beyond 33 missions in a day, it is highly unlikely and difficult to generate a valid test scenario. Each method had a parallel and sequential variation, and both Algorithm \ref{local_search_scheduling} and \ref{tabu_search_scheduling} used Algorithm \ref{schedule_initialize} for building a starting solution. For accurate comparison with the optimal, Gurobi was employed and tested upon the same data.

Based on the values shown in Figure \ref{fig_results_scheduling}, the two optimization algorithms had similar results. All were consistently under 8\% of the optimal, with some becoming quite close for the higher mission sets. That being said, the Tabu search did perform better than the Neighbourhood search in a very comparable time. The difference may not seem to be much based on the results in Table \ref{fig_results_table_scheduling}, though it should be noted that not a single case showed a better result for Neighbourhood search. Figure \ref{fig_times_table_scheduling} does display that it is technically faster, though at these speeds this is a negligible difference. Despite similar values, the consistency in the results show that the Tabu search is a better option. Admittedly, Figure \ref{fig_times_table_scheduling} does give the impression that upper and lower bounds are quite spread; however, the y-axis utilizes a very small incrementation. In actuality, they are fairly tight when further compared using Table \ref{fig_results_table_scheduling}. Additionally, total travel time is being compared instead of distance. This means the summed values are much smaller, explaining why it's easier to get within a better percentage of the optimal. 

The Gurobi optimizer almost always managed to achieve an optimal solution in a much slower time compared to the custom algorithms. The problem with scheduling is that it does not scale well, with larger missions counts making the space exponentially more difficult. While it was able to solve it in an acceptable time for missions below 24, anything above this became stuck. At points beyond 27 the gap between the upper and lower bounds stopped shrinking and the optimal was no longer improving. The optimizer was set to stop running after 20,000 seconds which is why Figure \ref{fig_results_time_scheduling} shows the last three values as being the same. This problem actually shows the issue with Gurobi in a case where the optimal needs to be determined quickly, as there is not a guarantee it will find it. This is not an unusal case in a disaster where timing can mean the difference between life and death. In this scenario these custom algorithms may be more reasonable, since in the case of 33 missions the Tabu search achieved a better lower bound than the value Gurobi had solved at that point in time. While in some cases it may be more useful to utilize the optimizer for a smaller count, the difficulty of this problem and the fact that the custom algorithms have a much faster runtime, does at least offer some advantages to alternatives.

Base on the results in both Figure \ref{fig_results_time_scheduling} and Table \ref{fig_times_table_scheduling}, the CUDA variations does not offer a better option at this scale. Both sequential versions are significantly faster in all cases. The problem with CUDA at this level is that the kernel call itself takes a lot of time to perform. It is possible to bundle everything into a single call, though it still will not be as fast as the other algorithms with these mission sizes. The CUDA implementation is about the same for both algorithms and still significantly faster than Gurobi. It is possible that at a higher mission count the parallel implementation may begin to outperform the sequential algorithms. Although, as previously mentioned it is unlikely to have more than 33 missions in a day for this set of data, meaning the only option would be if it was for scheduling a much longer period of time. In any case, while not an unreasonable result, at this scale the sequential Tabu search is still the best option.


\begin{table*}[]
\centering
\caption{Summary of Total Missions Times.}
\label{fig_times_table_scheduling}
\begin{tabular}{|l|l|l|l|l|l|}
\hline
\begin{tabular}[c]{@{}l@{}}\textbf{Number of}\\ \textbf{Missions}\end{tabular} & \begin{tabular}[c]{@{}l@{}}\textbf{Gurobi}\\ \textbf{Runtime}\end{tabular} & \begin{tabular}[c]{@{}l@{}}\textbf{Neighbourhood Search}\\ \textbf{Runtime}\end{tabular} & \begin{tabular}[c]{@{}l@{}}\textbf{Parallel Neighbourhood}\\ \textbf{Search Runtime}\end{tabular} & \begin{tabular}[c]{@{}l@{}}\textbf{Tabu Search}\\ \textbf{Runtime}\end{tabular} & \begin{tabular}[c]{@{}l@{}}\textbf{Parallel Tabu}\\ \textbf{Search Runtime}\end{tabular} \\ \hline
12                          & 1226                    & 0.0153                        & 0.548                                  & 0.0245                       & 0.578                                 \\ \hline
15                          & 2045                    & 0.0244                        & 0.532                                  & 0.0349                       & 0.615                                 \\ \hline
18                          & 2815                    & 0.0322                        & 0.544                                  & 0.0424                       & 0.575                                 \\ \hline
21                          & 5486                    & 0.0446                        & 0.519                                  & 0.0553                       & 0.507                                 \\ \hline
24                          & 9546                    & 0.0499                        & 0.611                                  & 0.0618                       & 0.642                                 \\ \hline
27                          & N/A                     & 0.0565                        & 0.618                                  & 0.0675                       & 0.561                                 \\ \hline
30                          & N/A                     & 0.0581                        & 0.599                                  & 0.0682                       & 0.645                                 \\ \hline
33                          & N/A                     & 0.0631                        & 0.686                                  & 0.0689                       & 0.611                                 \\ \hline
\end{tabular}
\end{table*}

\begin{table*}[]
\centering
\caption{Summary of Total Runtimes.}

\label{fig_results_table_scheduling}
\begin{tabular}{|l|l|l|l|}
\hline
\textbf{Number of Missions} & \textbf{Optimal Distance} & \textbf{Neighbourhood Search}                                                     & \textbf{Tabu Search}                                                      \\ \hline
12                          & 23.521                    & \begin{tabular}[c]{@{}l@{}}U: 25.495\\ L: 24.903\\ A: 25.199\end{tabular} & \begin{tabular}[c]{@{}l@{}}U: 25.374\\ L: 24.640\\ A: 25.007\end{tabular} \\ \hline
15                          & 24.348                    & \begin{tabular}[c]{@{}l@{}}U: 25.771\\ L: 25.119\\ A: 25.445\end{tabular} & \begin{tabular}[c]{@{}l@{}}U: 25.623\\ L: 24.858\\ A: 25.240\end{tabular} \\ \hline
18                          & 34.027                    & \begin{tabular}[c]{@{}l@{}}U: 36.834\\ L: 35.526\\ A: 36.008\end{tabular} & \begin{tabular}[c]{@{}l@{}}U: 36.138\\ L: 34.984\\ A: 35.611\end{tabular} \\ \hline
21                          & 46.853                    & \begin{tabular}[c]{@{}l@{}}U: 49.801\\ L: 48.029\\ A: 48.915\end{tabular} & \begin{tabular}[c]{@{}l@{}}U: 49.047\\ L: 47.380\\ A: 48.214\end{tabular} \\ \hline
24                          & 59.977                    & \begin{tabular}[c]{@{}l@{}}U: 62.365\\ L: 61.924\\ A: 62.144\end{tabular} & \begin{tabular}[c]{@{}l@{}}U: 61.793\\ L: 60.434\\ A: 61.164\end{tabular} \\ \hline
27                          & 57.841                    & \begin{tabular}[c]{@{}l@{}}U: 62.393\\ L: 58.950\\ A: 60.672\end{tabular} & \begin{tabular}[c]{@{}l@{}}U: 61.310\\ L: 58.859\\ A: 59.934\end{tabular} \\ \hline
30                          & 60.428                    & \begin{tabular}[c]{@{}l@{}}U: 63.904\\ L: 61.226\\ A: 62.565\end{tabular} & \begin{tabular}[c]{@{}l@{}}U: 62.379\\ L: 60.532\\ A: 61.456\end{tabular} \\ \hline
33                          & 66.564                    & \begin{tabular}[c]{@{}l@{}}U: 68.221\\ L: 66.956\\ A: 67.868\end{tabular} & \begin{tabular}[c]{@{}l@{}}U: 67.947\\ L: 66.425\\ A: 67.638\end{tabular} \\ \hline
\end{tabular}
\end{table*}

\clearpage

\section{Conclusion}
\label{sec:conclusion}
The results demonstrate the convenience of these custom algorithmic solutions. In all cases, Gurobi achieved an optimal in an increased time. Though not the universal, this is not acceptable in a disaster situation where time is not a luxury and fleet reorganization is required. In these scenarios, algorithms approaching the optimal become all the more valuable. Both algorithms were quite close to the Gurobi solution, though the Tabu slightly outperformed in all cases. In the latter simulations the Tabu even outperformed the Gurobi which could not calculate an effective permutation before the time termination expired. While parallel optimization may be an option in a larger scheduling scenario, it was not an improvement in this case. The kernel call resulted in a bottleneck which hindered any speedup gain. This should not eliminate the possibility of using this technique, though the situation should be considered. Overall, these algorithms are able to achieve acceptable results in an excellent time. They are adaptable to quick changes and can be run for multiple instances, displaying the effectiveness of this for the domain.


\bibliographystyle{unsrt}  
\bibliography{references}

\end{document}